\pdfoutput=1
\documentclass[11pt]{article}

\usepackage[final]{acl}

\usepackage{times}
\usepackage{latexsym}
\usepackage[ruled,linesnumbered]{algorithm2e}
\usepackage{xcolor}
\usepackage{tabularx}
\usepackage{amsmath}
\usepackage{graphicx}
\usepackage{hyperref}
\usepackage{booktabs}
\usepackage{enumitem}
\usepackage[symbol]{footmisc}

\usepackage{amsfonts} % mz: mathbb

\usepackage[T1]{fontenc}
\usepackage[utf8]{inputenc}

\usepackage{microtype}

\newcolumntype{Y}{>{\centering\arraybackslash}X}

\newtoggle{final}
\toggletrue{final}
\togglefalse{final} % comment this line for final version

\title{Automatic Unit Test Data Generation and Actor-Critic Reinforcement Learning for Code Synthesis}

\author{Philip John Gorinski\textsuperscript{*1}, Matthieu Zimmer\textsuperscript{*1}, Gerasimos Lampouras\textsuperscript{1} \\ \textbf{Derrick Goh Xin Deik\textsuperscript{2}, Ignacio Iacobacci\textsuperscript{1}}\\
Huawei Noah's Ark Lab, London\\
\textsuperscript{1}first.\{middle.\}last@huawei.com \textsuperscript{2}derrickg.deik@huawei.com
}

\begin{document}
\maketitle
\setlength\abovedisplayskip{1ex}\setlength\belowdisplayskip{1ex}
\begin{abstract}
The advent of large pre-trained language models in the domain of Code Synthesis has shown remarkable performance on various benchmarks, treating the problem of Code Generation in a fashion similar to Natural Language Generation, trained with a Language Modelling (LM) objective. In addition, the property of programming language code being precisely evaluable with respect to its semantics -- through the use of Unit Tests to check its functional correctness -- lends itself to using Reinforcement Learning (RL) as a further training paradigm. Previous work has shown that RL can be applied as such to improve models' coding capabilities; however, such RL-based methods rely on a reward signal based on defined Unit Tests, which are much harder to obtain compared to the huge crawled code datasets used in LM objectives. In this work, we present a novel approach to automatically obtain data consisting of function signatures and associated Unit Tests, suitable for RL training of Code Synthesis models. We also introduce a straightforward, simple yet effective Actor-Critic RL training scheme and show that it, in conjunction with automatically generated training data, leads to improvement of a pre-trained code language model's performance by up to 9.9\% improvement over the original underlying code synthesis LM, and up to 4.3\% over RL-based models trained with standard PPO or CodeRL.
\end{abstract}

\section{Introduction}
\renewcommand{\thefootnote}{\fnsymbol{footnote}}
\footnotetext[1]{Equal contribution}
\renewcommand{\thefootnote}{\arabic{footnote}}
Large Language Models (LLMs) have been dominating the field of Natural Language Processing (NLP) since the introduction of the transformer architecture \citep{vaswani2017attention} and the first large encoder-based \citep[BERT]{devlin2019bert}, decoder-based \citep[GPT]{radford2018improving, radford2019language}, and full transformer \citep[T5]{raffel2020exploring} pre-trained models. 
More recently, large transformer-based models began to expand beyond natural languages, most notably to the area of programming languages, where they are being applied to tasks such as Code Understanding and Code Synthesis (also referred to as Text-to-Code Generation), or Code Translation \citep{lu2021codexglue,wang2021codet5,tipirneni2022structcoder}.
Often, models addressing specific tasks build upon large pre-trained (code) language models, i.e. ``Foundation Models'' \citep{bommasani2021opportunities}, and apply further fine-tuning on task-specific data under a fully supervised LM or Imitation Learning paradigm.

For Code Synthesis, that is, the generation of programming language code conditioned on a natural language prompt, Reinforcement Learning (RL) has recently gained traction as an alternative or complementary training method \citep{le2022coderl,wang2022compilable}. These approaches make use of the fact that, as opposed to natural language, correctness of code is relatively straight forward to evaluate through compilation (syntax) and Unit Tests/functional testing (semantics).

\begin{figure}[t]
    \centering
    \includegraphics[width=\columnwidth]{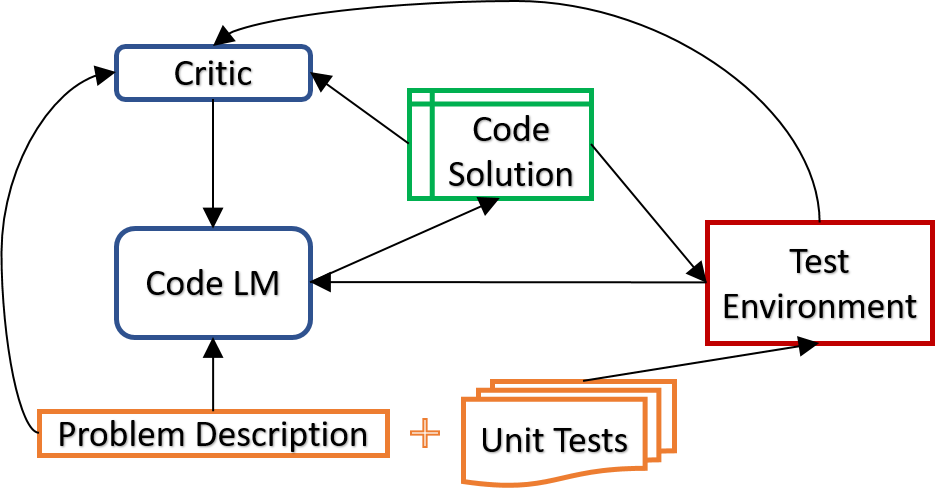}
    \caption{Diagram of overall Actor-Critic RL approach.}
    \label{fig:approach}
\vspace{-2mm}
\end{figure}

However, RL approaches that train on code functionality rely on data with corresponding Unit Tests, that are often only available in very limited quantities or not at all. This is in stark contrast to approaches that employ Language Modelling objectives, trainable on large quantities of crawled data. Overcoming this lack of Unit Test paired data has the potential to greatly improve the results achievable with such methods.

In this work, we also employ Reinforcement Learning to improve the performance of a pre-trained code language model on the Code Synthesis task. Modelling the reward based on Unit Test pass rates, we introduce a simple yet effective RL approach to fine-tuning Code Synthesis models, based on Policy Gradients and a simple feed-forward Critic model. In an effort to overcome the data sparseness issues of previous approaches, we introduce a heuristic approach to generate large data of natural language problems, function signatures, and Unit Tests, and show that the automatically generated data can help further improve Code Synthesis performance. Finally, we release our code, models, new dataset of problems, signatures, and Unit Tests.\footnote{Code and data available at \url{https://github.com/huawei-noah/noah-research/tree/master/NLP/PanguCodeRL}}

\section{Related Work}

The advent of large language models for programming languages was initially driven by repurposing general natural language understanding (NLU) and generation (NLG) approaches to the new domain. Models such as CuBERT \cite{kanade2020pretrained}, CodeGPT \cite{codeXGlue2021}, and CodeT5 \cite{wang2021codet5} greatly reused the natural language models, but trained on source code to provide them skills in the new context. 

\subsection{Large Language Models for code synthesis}

A number of LLMs have been focused on the task of text-to-code generation. \citet{codex} introduced CodeX, decoder-only language models trained on publicly available code from GitHub.
\citet{alpha_code} introduced AlphaCode, a set of sequence-to-sequence models trained on programming competition data
and GitHub code in several programming languages; they improved performance by over-generating and pruning solutions.

CodeGen~\cite{nijkamp2022conversational} proposed to tackle the code synthesis task through user conversations, i.e., the possibility of doing code generation and refinement from follow-up interactions. 
TiCoder \citet{lahiri2022interactive} followed a similar path; their approach generates many diverse solutions like AlphaCode, but also leverages user feedback by generating tests that can discriminate between proposed solutions. Finally, CodeT \citep{chen2023codet} proposed a unified model that generates both code snippets and Unit Tests that matched the program descriptions and were used to filter out some of the solutions. 

As is clear from the recap above, proposed models have been evolving over time, taking into consideration the signal of code behaviour to improve their performance.

\subsection{RL for Code}

One feature of programming language code is that can be precisely evaluated, and some research proposed to leverage Reinforcement Leaning
using code feedback as a reliable reward signal during training.
CompCoder, introduced by \citet{wang2022compilable}, proposed, among other things, to generate code in an RL environment, and reward those snippets which were able to compile, while punishing the others. CodeRL \cite{le2022coderl} went one step further, and proposed to reward code generation when the snippets were able to pass Unit Tests used to define the code behaviour.

Although our work is similar to the latter approach, we have fundamental differences: (i) our model relies exclusively on RL while in CodeRL, the critic is also used during inference; (ii) our critic is much simpler -– a simple FFNN on top of the PLM (c.f. Section~\ref{sec:rl}); (iii) our critic is not frozen during RL training (c.f. Equation~\ref{eqn:critic}), keeping it aligned with the policy; (iv) ours is trained on the expected reward including the policy KL term (c.f. Equation~\ref{eqn:kl}), which is impossible for CodeRL's as KL depends on the current policy; (v) ours gives a single score to a solution, similar to its training objective (c.f. Equations~\ref{eqn:critic_score} -- \ref{eqn:critic}), while CodeRL scores all tokens which is not fully aligned with its training; (vi) CodeRL as well as our approach require Unit Tests for training. We propose a way to create augmentation data of new pairs of code description and tests (c.f. Section~\ref{sec:ut}) which are then used to improve the capabilities of code synthesis.

\subsection{Code Datasets with Functional Tests}
As opposed to language modelling, which only requires continuous data for training, Reinforcement Learning as described above requires Unit Tests to model the reward.
Some code datasets have been released that contain code along with associated tests, such as MBPP \citep{austin2021program} and MTPB \citep{nijkamp2022codegen}; however these are typically very small, in the region of a few hundred training examples, therefore containing very few problems and offering limited learnable information in the context of large code synthesis models.

APPS \citep{hendrycks2021measuring} constitutes a larger code dataset of 10,000 total instances, 5k of which are reserved for training, crawled from various code challenge sites. While it can be a valuable resource, APPS is not built with function-level code and tests in mind, but contains larger scripts that expect to read inputs from standard-in, and write outputs to standard-out as strings.
This is not desirable for the models presented here, which are trained to synthesise function-level code, for well-defined and typed inputs and outputs. Converting APPS to a training format compatible with this objective is likely possible, but would require a considerable engineering or human effort.

\section{Approach}
\subsection{Task Definition}
We address the task of Code Synthesis, that is, the generation of Programming Language code conditioned on a Natural Language problem description, with Reinforcement Learning.
During RL training and evaluation, we employ Unit Tests, which allow for semantic evaluation of function level code, that is, they precisely define the expected behaviour of the code in terms of computation results.\footnote{Strictly speaking, precise evaluation requires Unit Tests that cover any and all potential edge cases for this work, we make the simplified assumption that a subset of all potential tests is sufficient to evaluate generated code.}

Compared to Natural Language, for which such precise and large scale semantic evaluation is notoriously costly and challenging \cite{sai2022survey}, the nature of programming language code, which is by definition executable and has formal syntax and semantics, makes it relatively straight forward to check for correctness \cite{allamanis2018survey}; this property makes it well suited to the application of Reinforcement Learning.

Owing to the Unit Test-based approach to training, and follow previous work \citep{codex}, we condition generation on the function signature, so as to guarantee the Unit Tests can be applied to the generated code, and provide a more reliable training signal.\footnote{Alternatively, we could massively over-generate code for each NL input, and only keep those generated functions compatible with the Unit Tests, or assign negative rewards to incompatible code during RL training. While this constitutes interesting future work, since function signatures are available through target Unit Tests anyway, we chose to include signatures in the generation condition.}
The natural language problem formulation and function signature combined constitute the \emph{prompt}, and the objective is for a model $M$ to generate code that accurately captures the described problem. An example is shown in Figure~\ref{fig:prompt}.

\begin{figure}[t]
    {\small
    \begin{tabularx}{\columnwidth}{lX}
         \textbf{NL Problem:} & Write a function that calculates the n-th Fibonacci number. \\
         \textbf{Signature:} & \texttt{def fib(n):} \\
         \textbf{Code:} & \hspace{1em}\texttt{if n == 0:}\\
         & \hspace{2em}\texttt{return 0}\\
         & \hspace{1em}\texttt{if n == 1:}\\
         & \hspace{2em}\texttt{return 1}\\
         & \hspace{1em}\texttt{return fib(n-2)+fib(n-1)} \\
         \textbf{Unit Tests:} & \texttt{fib(0)==0}; \texttt{fib(12)==144}; \texttt{fib(8)==21}
    \end{tabularx}
    }
    \caption{Example of Natural Language problem and code signature, together forming the code synthesis model prompt, and associated example code.}
    \label{fig:prompt}
\end{figure}

In this work, we specifically consider the synthesis of function-level code in the \emph{Python} programming language. However, the general approaches to RL training (Section~\ref{sec:rl}) and obtaining Unit Tests (\ref{sec:ut}) should be applicable to other settings, e.g., class- or project-level synthesis, and target programming languages with small modifications.

\subsection{Actor-Critic RL for Code Synthesis}
\label{sec:rl}

\paragraph{Reinforcement Learning} As a general ML approach, RL has been an active area of research for a long time \citep{kaelbling1996reinforcement,sutton2018reinforcement}. Compared to other methods such as Language Modelling, it does not rely on available gold-standard data -- though in settings with Imitation Learning, such data can additionally be used -- but instead makes use of a learning environment that assesses the model's solution(s), and gives feedback in the form of a reward signal, an indicator of how good the proposed solution worked. The training objective is to maximise the reward, i.e., the goodness of the model's output wrt. some assessment function.

Formally, an RL problem can be described as a Markov Decision Process (MDP, \citet{mdp}), a tuple $\langle S,A,P,R\rangle$ where $S=\{s_1,\ldots,s_n\}$ is a set of states, $A=\{a_1,\ldots,a_m\}$ a set of (possibly restricted) actions, $P=\{P(s_{t+1}| s_t, a)\}$ gives the probabilities of transitioning from state $s_t$ to another state $s_{t+1}$ at time $t$ with action $a$, and $R=\{R(s, a)\}$ gives the immediate rewards assigned to the state $s$ with the action $a$.
Many RL approaches have been proposed over the years, and a full review exceeds the scope of this paper. For an overview of available model-free and model-based RL algorithms and their applications, we refer the reader to \citet{sutton2018reinforcement}.

For the task of code synthesis, in general, the code language model serves as the RL actor's policy. The states $s_t$ are sequences given by the token-wise generated output up to time step $t$;
each action corresponds to the next token in the sequence.
In this setting, the transition probabilities are deterministic: the action is concatenated to the sequence $s_t$ to obtain $s_{t+1}$.
An episode ends when the end-of-sentence token is chosen or when a maximum length of sequence $T$ is reached.
The policy can select actions according to classic exploration strategies, e.g., it can act in an $\epsilon$-greedy fashion by selecting the next output token with highest probability, sample from the distribution, or even select a random token of the model's output vocabulary.

Considering the size of modern (code) generation LMs, with output vocabularies in the tens of thousands, as well as the length of sequences to generate which can be hundreds of tokens, it becomes obvious that training the policy ``from scratch'', i.e., starting with a randomly initialised policy LM and randomly exploring trajectories, is infeasible for the given task. Similarly, the strategy of selecting output tokens completely at random during generation is usually discarded, as most of the many available actions -- that is, tokens -- would simply lead to syntactically invalid code.

Instead, it is common practice to initialise the policy network with a pre-trained code language model, typically obtained by training with a language modelling objective on very large datasets. This ensures that the distributions over states and actions are already reflecting some likely ``good'' tokens, that is, the policy on its own is already able to generate \emph{syntactically} correct code. The goal when applying RL then becomes to update the LM parameters towards generating \emph{semantically} correct code, i.e., code that not only compiles and runs, but produces the desired output.

\paragraph{Reward Function} Code Synthesis reward can be modelled in various ways. \citet{wang2022compilable} checks for compilability of generated code. \citet{le2022coderl} additionally considers the outcome of Unit Tests, that is, whether the generated code \emph{behaves} as expected, and assign $-1$ if the code does not compile, $-0.6$ if it cannot be executed, $-0.3$ if it fails any Unit Tests, and $+1$ if it passes them all.

The present approach also considers Unit Tests, however, we want to reflect the intuition that \emph{partially} correct functions should also receive some reward distinguishing them from completely non-functioning code. As such, our reward function is composed of a first sparse part $R_\text{f}(s_t)$:
\begin{equation}
R_\text{f}(s_t) = 
\begin{cases}
    \lambda*\left(\frac{\text{\# passed UTs}}{\text{\# all UTs}}\right)^{\eta} & \text{\small if $s_t$ compiles} \\
    -10 & \text{\small otherwise}
    \end{cases}
\end{equation}
where the hyper-parameter $\lambda$ is a reward scaling factor, and $\eta$ controls the impact of the first passing Unit Tests; typically, we want to reward a solution for passing from zero to one valid Unit Test more than passing from one valid test to full correctness.

To avoid the model from over-fitting and moving too far away from the original PLM $\pi_0$, our complete reward function is given as:
\begin{multline}
R(s_t, a_t)=
    R_\text{f}(s_t) - \\
    \zeta \text{KL}\big(\pi_\theta(a_t|s_t) || \pi_0(a_t|s_t)\big)
\label{eqn:kl}
\end{multline}
where $\zeta$ is automatically adjusted to verify the constraint $\mathbb{E}[\text{KL}\big(\pi_\theta(a_t|s_t) || \pi_0(a_t|s_t)\big)] \leq \rho $ by the approach proposed in \citet{ziegler2020finetuning} where $\rho$ is a hyperparameter controlling how far the policy can move away from the original PLM.

One challenge arising from the Code Synthesis setting is the lack of immediate rewards required in the original RL formulation since code can only be evaluated for functional correctness once it is complete.
This means that during RL training, we have to wait for full sequences $\langle a_1 \ldots a_T \rangle$ of realised actions in order to assign a reward $R_f$ to the final generated token. This results in a reward for the overall generated code, but, the assignment of step-wise rewards is not straight forward to achieve.

\citet{le2022coderl} proposed to pretrain another external model to estimate step-wise rewards. However,
training such a step-wise reward predictor is in itself a very challenging task, which we will leave to future work. We found that using atomic rewards with REINFORCE works well in the proposed setting, in combination with a Critic model that can help derive a more stable policy gradient.

\paragraph{Critic Model for Reward Baseline}
As mentioned above, the proposed original reward function for this work relies on Unit Tests to assess generated the code's functional correctness. Final test-based rewards can be propagated across all sequence time steps to update the policy LM.

In addition to the Unit Tests, we find that the application of a critic model that evaluates the generated code can further improve training. Intuitively, we might expect such a critic to better distinguish the ``quality'' of various pieces of code, even if they receive the same reward by the Unit Test-based reward function. For example, consider the code to generate the n-th Fibonacci number in Figure~\ref{fig:prompt}, and the alternative implementation below in Figure~\ref{fig:alt}.
\begin{figure}[h!]
\small{
\hspace*{3em}\texttt{def fib(n):}\\
\hspace*{5em}\texttt{if n <= 0:}\\
\hspace*{7em}\texttt{return 0}\\
\hspace*{5em}\texttt{if n == 1:}\\
\hspace*{7em}\texttt{return 1}\\
\hspace*{5em}\texttt{return fib(n-2) + fib(n-1)}
}
\caption{Alternative code for n-th Fibonacci number.}
\label{fig:alt}
\end{figure}

Both solutions differ only in a single token, where the first code checks if \texttt{n == 0}, and the alternative checks for \texttt{n <= 0}. When we consider the Unit Tests in Figure~\ref{fig:prompt} and calculate rewards according to our proposed reward function, both solutions would receive full rewards. However, it could be argued that the second code snippet is better than the first, which would throw an exception for inputs smaller than zero.\footnote{One could argue that code which throws a specific exception here would be even better, as strictly speaking, the standard Fibonacci sequence is undefined for negative values.}

Given enough time and varied Unit Tests for observed problems, we could expect a deep learning-based critic to pick up on such or similar nuances across code solutions, potentially allowing for a more fine-grained reward to be used in policy updates. As such, for this work, we propose an Actor-Critic approach to improving code synthesis.
The critic can be trained in parallel with the policy, updating its parameters on generated solutions and observed associated rewards.

While the emergent property of a critic being able to distinguish ``good'' from ``bad'' correct code is hypothetical, we give some insights into observations in that direction in Appendix~\ref{app:critic}

\paragraph{REINFORCE with Critic for Code Synthesis}

We provide a pre-trained code synthesis model, used as the RL policy, a set of natural language problems, function signatures, and initial solutions akin to Figure~\ref{fig:prompt}. Training hyper parameters include maximum epochs, and the number of solutions to generate for each problem per iteration.
Our approach uses a replay buffer that stores the original solutions found in the training data, as well as all valid solutions found during training.

In each training iteration, we first generate a number of solutions for each problem in the training set, and receive the associated rewards from the environment. Additionally, an equal number of known valid solutions is sampled from the replay buffer, and assigned the maximum reward automatically. Generated and sampled solutions from the training set for the current epoch; newly found valid solutions are stored in the replay buffer.

The algorithm then proceeds to update the policy for a number of steps, determined by the provided minibatch size. For each solution in the minibatch, the critic model predicts a critic score $V_{\omega}$, which is used in combination with the environment reward to compute the update score, and update the policy parameters $\theta$ with REINFORCE:
\begin{equation}
\hat{A}_i = r_i - V_{\omega}(q_i, \sigma_i, \mu^{\prime}_i)
\label{eqn:critic_score}
\end{equation}
\begin{equation}
\theta \leftarrow \theta + \alpha\nabla_{\theta}\sum_i \log \pi_{\theta}(\mu^{\prime}_i|q_i,\sigma_i
)\hat{A}_i
\end{equation}

Finally, we update the critic model parameters $\omega$ based on the same epoch training data. The critic is trained as a regression model, using the Mean Squared Error loss between the observed reward and the critic's prediction:
\begin{equation}
\omega\leftarrow\omega-\beta\nabla_{\omega}\sum_{i}\text{MSE}(r_i, V_{\omega}(q_i, \sigma_i, \mu^{\prime}_i))
\label{eqn:critic}
\end{equation}

\noindent The proposed approach is agnostic to the precise architecture of the policy language model and critic network; we show the architectures used for this work in Section~\ref{sec:experiments}.

Note that, as opposed to other work such as \citet{le2022coderl}, the critic here is not employed to predict step-wise rewards \emph{independently} of the current policy, but rather to reduce the variance of the policy gradient by introducing bias with a baseline as it is more standard in RL.
While it could be expected that step-wise estimates might be beneficial, we did not observe this in preliminary experiments where critics were trained on pooled representations of the full sequence (or on the final end-of-sequence token), but applied to hidden representations of each individual time step when estimating rewards. In many cases, using this way of step-wise reward estimation even lead to detrimental training outcomes. We attribute this behaviour of step-wise critics in our setup to the mismatch between critic training and inference stages, and instead opt to just use the critic to provide scores for the full sequence, reflecting its own training objective.

\begin{algorithm*}[ht!]
\small
\DontPrintSemicolon
\caption{Practical REINFORCE-based RL with Critic for Code Synthesis}
\label{algo:rl}
\SetKwInOut{Input}{input}\SetKwInOut{Output}{output}
\Input{\hspace{1em}Pre-trained model parameters $\theta_{0}$; set $B=\{(q_i, \sigma_i, \mu_i)\}_{i=1\ldots m}$ of $m$ NL problems, PL signatures, and initial solutions; number of max training epochs $N$; number of generated solutions per problem $n_{gen}$; mini-batch size $n_{mini}$; policy learning rate $\alpha$; critic LR $\beta$}
\Output{\hspace{1em}Updated model parameters $\theta '$}
\BlankLine
Initialise policy parameters $\theta\leftarrow\theta_{0}$\;
Initialise critic parameters $\omega$ randomly\;
Initialise replay buffer of valid solutions $B_{valid} \leftarrow B$\;
\For{$epoch=1$ \KwTo $N$} {
Initialise training buffer $B_{train}\leftarrow \emptyset$\;
    \For(\hspace{2em}\texttt{// generate \& get rewards; prepare train set}){$i=1\ldots m$} {
        Sample $n_{gen}$ solutions $\{\mu^{\prime}_{i,j}\}_{j=1\ldots n_{gen}}$ for $(q_i, \sigma_i)$ with $\theta$ from distribution $\pi$\;
        Compute rewards $r_{i,j}$ for each $\mu^{'}_{i,j}$ from environment\;
        Sample $n_{gen}$ valid solutions $\{\mu^{*}_{i,j}\}$ for $(q_i, \sigma_i)$ from $B_{valid}$, assign maximum reward $r_{max}$\;
        $B_{train}\leftarrow B_{train}\cup \{(q_i, \sigma_i, \mu^{\prime}_{i,j}, r_{i,j})\}\cup  \{(q_i, \sigma_i, \mu^{*}_{i,j}, r_{max})\}$\;
        \lForEach{$\mu^{\prime}_{i,j}:r_{i,j}=r_{max}$}{$B_{valid}\leftarrow B_{valid}\cup \{(q_i, \sigma_i, \mu^{\prime}_{i,j})\}$}
    }
    \For(\hspace{2em}\texttt{// policy update}){$k=1\ldots|B_{train}|/n_{mini}$} {
        Sample minibatch $\{(q_i, \sigma_i, \mu^{\prime}_{i}, r_{i})\}_{i=1\ldots n_{mini}}$ from $B_{train}$\;
        Compute $\hat{A}_i=r_i - V_{\omega}(q_i, \sigma_i, \mu^{\prime}_i)$ from reward and the critic\;
        $\theta\leftarrow\theta+\alpha\nabla_{\theta}\sum_{i}log\ \pi_{\theta}(\mu^{\prime}_i|q_i, \sigma_i)\hat{A}_i$\;
    }
    \For(\hspace{2em}\texttt{// critic update}){$k=1\ldots|B_{train}|/n_{mini}$} {
        Sample minibatch $\{(q_i, \sigma_i, \mu^{\prime}_{i}, r_{i})\}_{i=1\ldots n_{mini}}$ from $B_{train}$\;
        $\omega\leftarrow\omega-\beta\nabla_{\omega}\sum_{i}\text{MSE}(r_i, V_{\omega}(q_i, \sigma_i, \mu^{\prime}_i))$\;
    }
}
\end{algorithm*}

In terms of novelty, we introduce the usage of a replay buffer with abstract syntax trees filtering to significantly improve diversity and show that one can update the critic during training, instead of only pretraining it, leading to stable training and improved performance.
Algorithm~\ref{algo:rl} shows pseudo-code for our RL approach to Code Synthesis.

\subsection{Automatic Unit Test Dataset Generation}
\label{sec:ut}

\begin{figure*}[!ht]
    \centering
    \includegraphics[width=.8\textwidth]{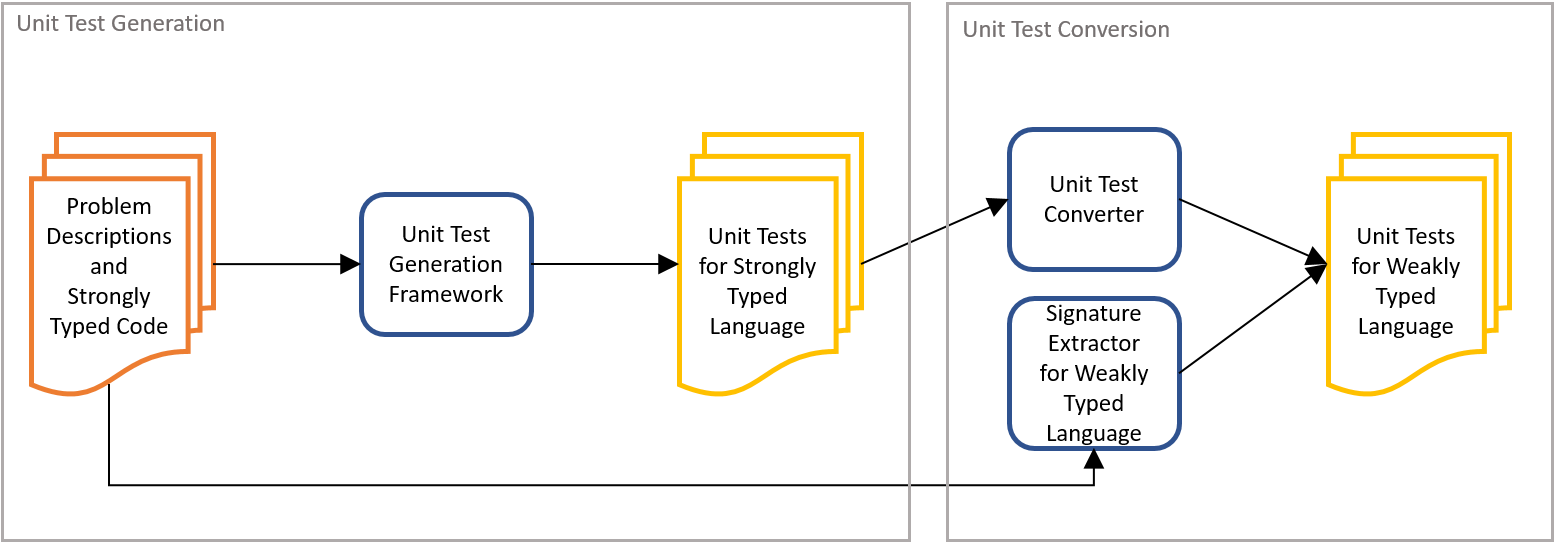}
    \caption{Overall outline of Unit Test generation and conversion approach.}
    \label{fig:utgen}
\end{figure*}

As mentioned above, existing code datasets with Unit Tests are often very limited in size, or do not target function-level Code Synthesis and evaluation. We therefore propose to employ automatic Unit Test generation to obtain tests for (essentially) arbitrary amounts of code.

Large code datasets, such as the Pile \citep{pile} have recently become readily available, providing hundreds of thousands if not more examples of code, in many programming languages. Such data is ideal for use in (pre-)training of large code language models with classic Language Modelling objectives. However, Unit Tests are not usually available for the code provided, or where they are, they might be provided indirectly as part of the code description or in other implicit ways, making them hard to extract.
On the other hand, the sheer amount of code available online makes it possible to try and use existing Unit Test generation approaches and, even if complete test suites can be generated for only a fraction of tried code, generate large quantities of code-test pairs.
Thus, we employ the following steps to obtain large amounts of Unit Test data for our task and RL approach:
{\begin{enumerate}[leftmargin=*]
    \setlength{\itemsep}{0pt}
    \setlength{\parskip}{0pt}
    \setlength{\parsep}{0pt}
    \item From a large code dataset such as the Pile, extract a large amount of function-level instances.
    \item Split each obtained instance into descriptions (e.g., doc strings, function comments) and code.
    \item Use an existing Unit Test generation framework to obtain Unit Tests for each instance's code.
    \item If Unit Test generation is successful, store the corresponding description, function signature, extracted code, and Unit Tests as an instance of the new training data.
\end{enumerate}
}

\noindent Depending on the chosen target synthesis language -- here, Python -- and Unit Test generation framework, this method might yield another complication. In particular, test generation for weakly typed languages such as Python or JavaScript is decidedly harder than for strongly typed languages like Java or C\#.
Even if the target language is weakly typed, we can mitigate this by running Unit Test above extraction and generation approach explicitly for some other, strongly typed language or languages. In such cases, we need to run include two additional steps into above procedure:
{\begin{enumerate}[leftmargin=*]
    \setlength{\itemsep}{0pt}
    \setlength{\parskip}{0pt}
    \setlength{\parsep}{0pt}
    \item[5.] Convert the original source signature into a signature compatible with the target language.
    \item[6.] Convert the generated Unit Tests to test statements compatible with the target language.
\end{enumerate}
}

\noindent The outline of this overall procedure is depicted in Figure~\ref{fig:utgen}; an illustration of Unit Test and signature conversion is in Appendix~\ref{app:conv} Figure~\ref{fig:conv}, and some examples of automatically generated augmentation instances used for training are given in Appendix~\ref{app:conv} Figure~\ref{fig:data}.
Of course, while we convert signatures and tests, this leaves the original code untranslated. While code translation would ideally be carried out as well, this task is an active area of research in itself \citep{roziere2020unsupervised,wang2021codet5,codeXGlue2021}, and incorporating it here would go beyond the scope of this work.

However, for our Reinforcement Learning approach, only the description, signature, and tests are strictly required: during training, the policy model is updated based on its own generated and evaluated code. In cases where we make use of automatically extracted code and Unit Tests, but have to drop the original code, we simply leave the initial replay buffer of the corresponding valid solutions in Algorithm~\ref{algo:rl} empty, and populate it over time with found valid solutions.

\section{Experiments}
\label{sec:experiments}
\paragraph{Setting} With our overall procedures to RL for Code Synthesis and Unit Test generation in place, we run a battery of experiments to evaluate the efficacy of our approach.

As a general dataset for the task, we use the Mostly Basic Python Problems Dataset (MBPP, \citealp{austin2021program}). MBPP provides a total of 964 instances of natural language problem description, an associated gold solution, and Unit Tests. We follow the official train and validation splits of 374 and 90 instances respectively, and combine the original few-shot prompting and test splits into a test set totalling 500 examples.

For all experiments, we use the pre-trained PanGu-Coder \citep{christopoulou2022pangucoder} Code LM with 300M parameters to initialise the policy.
We initialise the critic network on top of a copy of the same pre-trained model, as a two-layer MLP, taking as input the final representation of the end-of-sequence token of the underlying LM, downsampling to half the number of dimensions and again to a single output unit, with the ReLU activation function between layers. During critic training, we freeze the critic's copy of the code LM, and only update MLP.

We set the policy learning rate $\alpha=5e^{-7}$ and critic LR $\beta=1e^{-6}$. The maximum number of training epochs is 100, and we generate $n_{\text{gen}}=8$ solutions per problem per epoch. The reward hyper parameters are set to $\lambda=50$ and $\eta=0.5$. The target KL divergence is set to $\rho = 0.07$. When sampling solutions from the policy, we use nucleus sampling \citep{holtzman2019curious}, with $\text{p}=0.8$ and $\text{temperature}=0.95$.
Each training minibatch contains 32 solutions with a max length $T=512$.

For each model under consideration, we evaluate performance on the MBPP validation set after each epoch, and select the checkpoint with best validation score (based on greedy decoding) for test set evaluation. 
We report more details on the experiments and the baselines in Appendix~\ref{sec:impl_details}. Finally, ablation studies 
are carried out in Section~\ref{sec:ablation}.

\paragraph{Augmented Data} In order to augment our training data with additional, automatically generated Unit Tests, we randomly sample and extract an initial 100k function-level docstring-code pairs from a large Java dataset; crawled from existing, public GitHub repositories before May 2021. We only keep functions for which the accompanying description is in English as classified by the Lingua language detector,\footnote{\url{https://pemistahl.github.io/lingua-py/}} and is between 10 and 512 white-spaced tokens in length. We use the EvoSuite framework to automatically generate Unit Tests,\footnote{\url{https://www.evosuite.org/}} and discard instances with fewer than 2 tests, as well as those for which all tests result in the same outcome, e.g., if all tested inputs to a function with boolean return value result in \texttt{False}. 
After all filtering steps, we obtain an augmentation dataset with an additional $5,572$ instances of problem description, python signature, and Unit Tests.
Finally, we use the pre-trained language model to further filter out instance for which no valid solution can be found in 100 samples. This leaves a final amount of 2,229 instances in our augmented dataset.

\paragraph{Models and Results} We evaluate various baselines as well as our model, with the same underlying large pre-trained code generation model (Org. PLM): (i) A fine-tuned version of the pre-trained LM, tuned with CLM objective on the MBPP training set. (ii) A model trained with Proximal Policy Optimization (PPO) \cite{DBLP:journals/corr/SchulmanWDRK17}. (iii) A model trained with the CodeRL approach of \citet{le2022coderl}, on based on the same PLM described here. (iv) Applying our proposed training from Algorithm~\ref{algo:rl} on the MBPP training set (``Ours''). (v) Applying our proposed approach to learn from the automatically generated data exclusively (``Ours+Aug''). (vi) Perform RL training with Algorithm~\ref{algo:rl} on the combined data from MBPP training and augmented training sets (``Ours+All''). (vii) Rebalance the two datasets by decreasing the weight of the augmented unit tests in the loss by 0.2 (``Ours+All (0.2)'').
The performances of the various models on the MBPP test set are summarised in Table~\ref{tab:mbpp}. We report scores for greedy decoding, as well as pass@k \citep{chen2021evaluating} for $k=1$, $10$ and $100$, estimated on 200 samples.

\begin{table}[!t]
    \small
    \centering
    \setlength{\tabcolsep}{.9mm}{
    \begin{tabularx}{\columnwidth}{l|YYYY}
        \toprule
         & Greedy & Pass@1 & Pass@10 & Pass@100 \\
         \midrule
        Org. PLM & 22.0 & 15.6 & 40.0 & 57.1 \\
        CLM & 23.2 & 21.3 & 37.8 & 47.9 \\
        PPO & 27.2 & 22.1 & 42.5 & 54.2 \\
        CodeRL & 25.2 & 24.2 & 43.6 & 56.2 \\
        Ours & \underline{28.6} & \textbf{25.5} & \textbf{45.9} & 58.7 \\
        \midrule
        Ours+Aug & 21.8 & 18.9 & 41.8 & 57.6 \\
        Ours+All & 25.8 & 23.6 & 45.6 & \underline{58.8}\\
        Ours+All ($0.2$) & \textbf{29.2} & \underline{24.5} & \underline{45.4} & \textbf{58.9}\\
        \bottomrule
    \end{tabularx}
    }
    \caption{MBPP test results. The best model per metric is marked \textbf{bold}, the second best is \underline{underlined}.}
    \label{tab:mbpp}
\end{table}

\paragraph{Discussion}

All trained models improve greedy, pass@1 and pass@10 performance over the original code LM. Interestingly, only two, our approach trained on MBPP, or on all available data, improve pass@100. We conjecture that, as we generate $8$ solutions per problem during training, this might strengthen the models on metrics around or below that pass@k rate.

Supervised training with the CLM objective leads to mostly small improvements, which could be explained by the relative difference between MBPP split sets, where the 374 training instances offer only a limited signal for fully supervised learning. While some knowledge gained in training transfers to the test problems, the gold-standard code in the training set likely does not cover enough phenomena in the test set.

We observe that across all metrics, methods using our Actor-Critic approach outperform the baselines when trained on MBPP, with the model trained based on MBPP alone achieves the top rank in pass@1, pass@10, and pass@100.

To explain why our approach is better than PPO, we conjecture that learning the necessary stepwise critic is difficult for this task with very sparse rewards.
Moreover, it is known that dropout affects PPO stability, as the importance sampling ratio becomes stochastic \citep{hausknecht2022consistent}.
We had to disable dropout for PPO to make it stable, which might impede its generalisation capabilities.

Training our method on the augmented data alone improves over the original PLM, but falls short compared to RL models trained on MBPP. However, when combining the augmented and MBPP train data, our method achieves greedy improvement of $7.2$ points over the PLM and $0.6$ points over the same model trained only on MBPP. This indicates that there is indeed a useful training signal present in the automatically generated data, and it can be leveraged to improve performance.

Overall, we find that both, our Actor-Critic training method, as well as our method of augmenting the available training data automatically with Unit Test generation and conversion can contribute positive impacts on code synthesis model performance.

\paragraph{Out-of-domain Performance} To better evaluate the models' generalisation capabilities, we run evaluations on the 163 instances of the HumanEval dataset \citep{chen2021evaluating}. For all models, we evaluate the same checkpoints as previously determined, and summarise the results in Table~\ref{tab:he}.

\begin{table}[t]
    \small
    \centering
    \setlength{\tabcolsep}{.9mm}{
    \begin{tabularx}{\columnwidth}{l|YYYY}
        \toprule
         & Greedy & Pass@1 & Pass@10 & Pass@100 \\
         \midrule
        Org. PLM & \textbf{23.1} & 15.8 & 28.1 & 38.5 \\
        CLM & 14.6 & 15.1 & 25.1 & 33.8 \\
        PPO & 19.5 & 15.9 & 26.5 & 36.2 \\
        CodeRL & 18.9 & 16.9 & 29.4 & 41.6 \\
        Ours & 20.1 & 17.4 & 30.0 & 40.2 \\
        \midrule
        Ours+Aug & \underline{21.3} & 17.2 & 28.8 & 39.4\\
        Ours+All & 18.9 & \underline{17.8} & \textbf{31.7} & \textbf{45.9} \\
        Ours+All ($0.2$) & 20.7 & \textbf{18.2} & \underline{31.1} & \underline{42.5} \\
        \bottomrule
    \end{tabularx}
    }
    \caption{HumanEval test results. The best model per metric is marked \textbf{bold}, the second best is \underline{underlined}.}
    \label{tab:he}
\end{table}

We find that models trained with our approach consistently outperform the other RL-trained as well as the CLM baselines in all evaluation settings. Interestingly, the original PLM performs best when generating greedy solutions, while \emph{all} MBPP-trained models suffer here. This is likely an artifact of overfitting during fine-tuning on MBPP; however, all models have learned from the data, and (with the exception of CLM) they outperform the base model on sampling-based metrics.

Overall on HE, the model trained with our proposed approach performs best already without use of augmented data, except for Pass@100 where it is beaten by the CodeRL-trained model. However, this is remedied by the use of augmentation data.

Interestingly, the effect of using the combined original and augmented data has provided a larger relative performance boost on the out-of-domain HE data when compared to the results on MBPP; on Pass@100, it can increase performance by up to $5.7\%$. However, these improvements have to be taken with a grain of salt, given the small number of test instances provided by HumanEval.

\section{Ablation Studies}
\label{sec:ablation}
In this section, we verify if several components of our method are necessary.
In those experiments, to reduce the needed computations, we only train on MBPP train without the augmented Unit Tests.
The validation process is kept the same.

\subsection*{Sensitivity to the target KL}

\begin{table}[ht!]
    \small
    \centering
    \setlength{\tabcolsep}{.9mm}{
    \begin{tabular}{l|cccc}
        \toprule
         & Greedy & Pass@1 & Pass@10 & Pass@100 \\
         \midrule
        Ours $\rho = \infty$ & \underline{27.8} & \underline{25.1} & 44.6 & 56.1 \\
        Ours $\rho = 0.1$ & 27.4 & 24.4 & \underline{45.6} & 58.4 \\
        Ours $\rho = 0.08$ & 27.6 & 24.5 & 44.9 & 57.4 \\
        Ours $\rho = 0.07$ & \textbf{28.6} & \textbf{25.5} & \textbf{45.9} & \underline{58.7} \\
        Ours $\rho = 0.05$ & 26.6 & 22.9 & 45.5 &  57.8 \\
        Ours $\rho = 0.02$ & 26.8 & 22.5 & 45.8 & \textbf{60.5} \\
        \bottomrule
    \end{tabular}
    }
    \caption{Evaluation results on the MBPP test dataset for different values of target KL divergences after training on MBPP train.}
    \label{tab:kl}
\end{table}

We show the impact of the target KL divergence $\rho$ in Table~\ref{tab:kl}.
We observed that $\rho = 0.07$ is a good trade-off over all criteria; it notably has a large impact on pass@10 (and ultimately in pass@100).
When setting $\rho = \infty$, we observed that the averaged KL divergence would not exceed $0.2$.

\subsection*{Replay buffer of valid solutions}

In this experiment, we check that using a replay buffer is helpful.
To do so, we use the same algorithm but without storing the valid solutions which reduces to REINFORCE with a critic.
Table~\ref{tab:rb} shows that having a replay buffer is especially crucial to maintain a good diversity in pass@10 and in pass@100.

\begin{table}[ht]
    \small
    \centering
    \setlength{\tabcolsep}{.8mm}{
    \begin{tabular}{l|cccc}
        \toprule
         & Greedy & Pass@1 & Pass@10 & Pass@100 \\
         \midrule
        Ours - replay buffer & 26.6 & 24.2 & 38.0 & 45.0 \\
        Ours & {\bf 27.8} & {\bf 25.1} & {\bf 44.6} & \textbf{56.1} \\
        \bottomrule
    \end{tabular}
    }
    \caption{Evaluation results on the MBPP test dataset with or without a replay buffer of valid solutions after training on MBPP train.
    No KL divergence was used.}
    \label{tab:rb}
\end{table}

\section{Conclusions and Future Work}
We have presented two novel ways of improving code synthesis models.
On the training side, we introduced a practical Reinforcement Learning approach, based on an simple Actor-Critic REINFORCE algorithm. This approach shown to be simple yet effective, improving over CLM as well as PPO-trained models.
To provide additional learning signals, we furthermore introduced a method of extracting function-level strongly-typed code from large crawled code datasets, automatically generating Unit Tests for extracted functions, and conversion for use with a weakly typed target language. The generated data was able to further improve model performance, getting the overall best results for greedy code generation.

In the future, we would like to explore our methods in settings with wider scope, i.e., go from function-level code to class- or even project-level. This poses significant challenges, as rewards will become even sparser, and data augmentation will have to take much larger scope into account. 

\section{Limitations}

We are aware of some limitations affecting this work. First, the choice of MBPP as the primary dataset for training, validation, and testing provides only a small amount of data, with a total of less than $1,000$ instances. As mentioned, datasets that are directly applicable to the task and approach here, that is, dataset that provides well-defined function-level code as well as Unit Tests to evaluate the code's functional correctness are not easily available in large quantities.

We worked towards alleviating this limitation with our data augmentation approach, however, this in itself is limited by various factors. We chose Java as the strongly typed source language, which follows a very different language paradigm than our target, Python, with its Class-first approach. As such, it took a large amount of data to get the initial Java functions, before even being able to generate Unit Tests. The approach is further limited by the speed of test generation, and generating the initial ${\sim}5,5k$ instances of the augmented data takes a considerable amount of time. The original generated UT data is also rather noisy, as evidenced by the filtering steps that had to be applied to get the final ${\sim}2,4k$ instances used for training. On the other hand, the process is automated, and can be run in the background to generate arbitrary amounts of data. While the first iteration of the data released with this work is thus relatively small, we plan to make more data available in subsequent releases.

We also try to alleviate the result's by additionally testing on HumanEval. This itself is a very small dataset with fewer than $200$ instances; however, as we do not use it for training, but only to evaluate the various baselines trained on MBPP and/or our augmented data, we think it provides valuable insights especially wrt. generalisation capabilities instiled into the models through the various training approaches.

The CodeRL baselines based on \citet{le2022coderl} uses the same underlying model as used for our other approaches. While it could be argued that their original CodeT5-based approach should be part of the evaluation, we believe there are good reasons to employ their \emph{method}, rather than their \emph{model}: First, the policy model employed by \citet{le2022coderl} is significantly larger and of a different architecture than our ${\sim}300M$ parameter GPT2-like PLM, making direct comparison difficult. However, their RL approach -- like ours -- should be independent of model architecture, so comparing it against our approach on top of the same model seems fair. Second, most of the improvement in their work seems to stem not from their approach to RL training, but from using a second trained critic at test time, which assesses generated code for likely ``break-down'' points, and re-starts synthesis from there to find a better solution. This, in effect, massively over-generates solutions to find a working one, whereas we do not employ any such code break-down and refinement at test time, but use the final policy model as-is. In fact, our experiments indicate that without such over-generation at test time, our RL approach is able to outperform CodeRL even without using any augmented data.

\bibliography{bibliography}
\bibliographystyle{acl_natbib}

\appendix

\section{Implementation details}
\label{sec:impl_details}

\subsection{Actor Network}
We use a GPT2-like code language model trained on a large code dataset with a Causal Language Modeling objective. The model has 25 transformer decoder blocks, with an embedding size of $1024$ parameters, and a total of $317$ million parameters (excluding word embeddings). The vocabulary embedding the question consists of $32k$ tokens and the one encoding the python code also consists of $32k$ tokens.

\subsection{Critic Network}
To optimise computations and memory, we build our critic on top of the original PLM.
Since we need to compute $\pi_0$ for every solution for the KL divergence part of the reward, we reuse the embedding of the last token as input of the critic.
The multilayer perceptron is composed of the following layers: a linear layer of 512 neurons, the hyperbolic tangent activation function and a final layer of 1 neuron.
The gradient is only used to update the last two linear layers without the original PLM.

\subsection{Replay Buffer}

For a given problem, many alternative solutions exist, notably by adding irrelevant and unused lines of code.
To avoid promoting those kinds of solutions, before adding a solution that obtained the maximum reward to the replay buffer of valid solutions we perform a reduction of the code by converting into an abstract syntax tree.
From there, we remove all unused functions, comments and lines outside the scope of the main function appearing after the main function.
After those simplifications, we also verify that the same solution does not already exist in the replay buffer.

\subsection{PPO}
The same actor architecture and the same critic architecture are used for the PPO baseline.
The main difference lies in the performed updates.
Instead of trying to predict the scalar outcome of a trajectory, the PPO critic is predicting a step-wise score given by Generalised Advantage Estimation (GAE) \citep{schulman2018highdimensional}.
The update of the actor is also different and relies on the step-wise score given by the critic.
We let the reader refer to \citet{DBLP:journals/corr/SchulmanWDRK17} for the exact loss.

For fair comparison, the reward function, minibatch size, number of generated trajectories per question and learning rates are kept the same as for our method.
For the additional hyperparameters introduced by PPO, we kept standard values: the GAE-$\lambda$ is 0.97, $\gamma$ is set to 0.99 and the clip of the importance sampling ratio is 0.1.

The KL divergence has also optimised and we kept the value $\rho = 0.07$.

\subsection{CodeRL}

For the CodeRL baseline, we reused the same actor and critic architecture, however, the critic is pretraining the full transformer and not only the last layers.
To perform the pretraining, as in \citet{le2022coderl}, we first train a policy exclusively with CLM on MBPP train.
We select the best checkpoint for this policy according to the validation performance over MBPP of the greedy decoded solution.
Then, we continuously generate solutions with that policy to pretrain the critic (including ground truth solutions).
We also selected the best checkpoint for this critic according to the mean square error in validation over MBPP.
Once the critic is pretrained, it is kept frozen during RL training.
During the critic pretraining, a max-pooling of the features is used in order to predict the outcome of a trajectory.

The actor is updated with the step-wise score given by the critic and the greedy baseline with a REINFORCE-like loss.

For fair comparison, the reward function, minibatch size, number of generated trajectories per question and learning rates are kept the same as for our method.

\subsection{Hardware Specifications}
We trained our models using a single GPU (Nvidia Tesla V100-SXM2 32GB) and an Intel(R) Xeon(R) Platinum 8160 CPU @ 2.10GHz CPU.
To train 30 epochs, around 24 hours were needed for each RL methods involving augmented Unit Tests (around 670k solutions were generated for Ours+All).
To test on MBPP, around 6 hours were needed to generate 100k solutions.

\section{More results}

In this section, we report additional results gathered during the experiment phases.

In Table~\ref{tab:valid}, we report the best validation for each method and the associated training pass@1 for the same checkpoint.

\begin{table}[h]
    \small
    \centering
    \begin{tabular}{l|cc}
        \toprule
         & Validation (greedy) & Training (pass@1)  \\
         \midrule
        CLM & 30.7 & 56.6 \\
        PPO & 31.1 & 43.5 \\
        CodeRL & \underline{34.4} & 57.8 \\
        Ours & {\bf 37.7} & 63.4 \\
        \midrule
        Ours+Aug & 29.3 & 67.4\\
        Ours+All & 33.6 & 79.9\\
        Ours+All (0.2) & 31.5 & 71.8\\
        \bottomrule
    \end{tabular}
    \caption{Best training and validation on MBPP for each method. The pass@1 was computed with the 8 generated trajectories. On the bottom part of the table the training performance is computed with the augmented unit tests exclusively for Ours+Aug and with MBBP train and the augmented unit tests for Ours+All.}
    \label{tab:valid}
\end{table}

\begin{table}[ht!]
   \small
   \centering
   \begin{tabular}{l|ccc}
       \toprule
        & Greedy \\
        \midrule
        Ours & 0.9\\
        Ours+Aug & 1.8\\
        Ours+All & {\bf 2.3}\\
       \bottomrule
   \end{tabular}
   \caption{Evaluation results on filtered augmented data, not encountered during training.}
   \label{tab:aut_test}
\end{table}

In Table~\ref{tab:aut_test}, we analyse the performance of our different methods on augmented Unit Tests not used for training; notably, these are data points from the original augmented dataset for which no valid solution could be found by the base PanGu-Coder PLM, among 100 samples. We also filtered questions and signatures such that they are unique in the testing set and do not appear in the training. 
This testing set is composed of $1,428$ instances.
Because of such a large number of instances, we only focused on greedy decoding evaluation.
On this test data, both our models trained on the augmented data (Ours+Aug), as well as on all available training data (Ours+All) are better than the model trained on MBPP alone, showing that learning on more UT increases the overall model performance.

\section{Observations of Critic Scores}
\label{app:critic}
As mentioned in a motivating example in Section~\ref{sec:rl}, we could reasonably expect a deep learning-based critic, such as used in this work, to develop some understanding of ``good'' and ``bad'' code. This would allow a critic to distinguish even working solutions (i.e., different solutions that pass all Unit Tests), assigning scores related to code quality.

While such emergent properties in neural critics are purely hypothetical on our end, we still are interested in whether we can identify patterns in critic score assignments.

To this end, we manually inspected $62$ problems from the HumanEval dataset for which our final model (``Ours+All'' in previous experiments) was able to generate multiple valid solutions, according to unit tests. We then assigned scores to each solution using the final critic, selecting the solutions with highest and lowest critic scores for comparison.
We established a number of observations:

\begin{itemize}
    \item On average, higher rated solutions were slightly longer than their lower rated counterparts, with an average length of 7.21 and 7.01 lines of code produced respectively. However, this is largely attributed to a single outlier for which the preferred solution had 43 lines, and
\end{itemize}
\clearpage
\begin{figure*}
    \begin{minipage}{.45\textwidth}
        \small{
            \hspace*{3em}\texttt{def fib(n: int):}\\
            \hspace*{5em}\texttt{if n == 0:}\\
            \hspace*{7em}\texttt{return 0}\\
            \hspace*{5em}\texttt{if n == 1:}\\
            \hspace*{7em}\texttt{return 1}\\
            \hspace*{3em}\texttt{return fib(n-2)+fib(n-1)}
        }
    \end{minipage}
    \begin{minipage}{.55\textwidth}
        \small{
            \hspace*{3em}\texttt{def fib(n):}\\
            \hspace*{5em}\texttt{if n <= 0:}\\
            \hspace*{7em}\texttt{return 0}\\
            \hspace*{5em}\texttt{if n == 1:}\\
            \hspace*{7em}\texttt{return 1}\\
            \hspace*{5em}\texttt{return fib(n-2) + fib(n-1)}
        }
    \end{minipage}
    \caption{Motivating (hand-written) examples of implementations for Fibonacci, repeated from Figures~\ref{fig:prompt} (left) and~\ref{fig:alt} (right)}
    \label{fig:fib_manual}
\end{figure*}

\begin{figure*}[ht!]
    \begin{minipage}{.45\textwidth}
    \small{
        \hspace*{3em}\texttt{def fib(n: int):}\\
        \hspace*{5em}\texttt{if n < 2:}\\
        \hspace*{7em}\texttt{return n}\\
        \hspace*{5em}\texttt{return fib(n-2) + fib(n-1)}\\
        \hspace*{1em}\\
        \hspace*{1em}\\
        \hspace*{1em}
    }
    \end{minipage}
    \begin{minipage}{.55\textwidth}
        \small{
            \hspace*{3em}\texttt{def fib(n: int):}\\
            \hspace*{5em}\texttt{if n < 2:}\\
            \hspace*{7em}\texttt{return n}\\
            \hspace*{5em}\texttt{prev, curr = 0, 1}\\
            \hspace*{5em}\texttt{for i in range(2, n+1):}\\
            \hspace*{7em}\texttt{prev, curr = curr, prev + curr}\\
            \hspace*{5em}\texttt{return curr}
        }
    \end{minipage}
    \caption{Highest-scored (left) and lowest-scored (right) working implementations of the Fibonacci sequence, as generated and scored by our final policy and critic models.}
    \label{fig:fib_sols}
\end{figure*}

\begin{figure*}[h!]
    \centering
    \includegraphics[width=\textwidth]{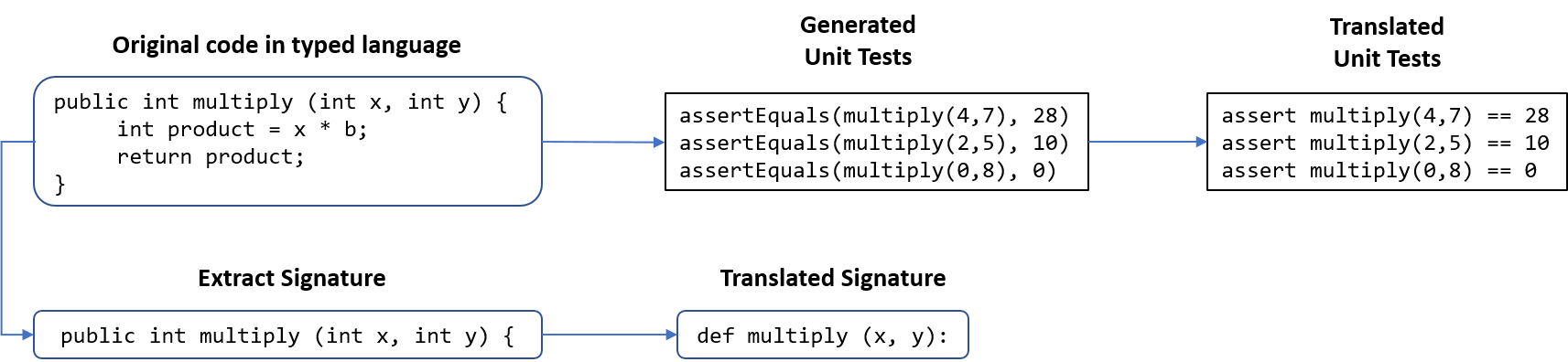}
    \caption{Example of Unit Test and function signature conversion}
    \label{fig:conv}
\end{figure*}

\begin{table*}[t]
    \centering
    \small{
    \begin{tabularx}{\textwidth}{lX}
         \textbf{Function Description:} & A method that return Yes only if both monkeys are smiling or not smiling @param aSmile @param bSmile @return yes only if both monkeys are smiling or not smiling , no if not TODO : Your code goes here \\
         \textbf{Python Signature:} & \texttt{def monkeyTrouble2(aSmile, bSmile):} \\
         \textbf{Unit Tests:} & \texttt{assert monkeyTrouble2(False, False) == "Yes"}\\
                              & \texttt{assert monkeyTrouble2(True, True) == "Yes"}\\
                              & \texttt{assert monkeyTrouble2(True, False) == "No"}\\
                              & \texttt{assert monkeyTrouble2(False, True) == "No"}\\
        \midrule
        \textbf{Function Description:} & Returns the max of a and b. Do not step into this function. This function may have a bug, but if it does, you should find it by stepping over, not into. \\
         \textbf{Python Signature:} & \texttt{def max(a, b):} \\
         \textbf{Unit Tests:} & \texttt{assert max(0, 581) == 581}\\
                              & \texttt{assert max(-1, -1) == -1}\\
                              & \texttt{assert max(581, 373) == 581}\\
                              & \texttt{assert max(0, 0) == 0}\\
        \midrule
        \textbf{Function Description} & this loop works by comparing the first character of string2 against the *characters of string1. If it isn't found in the first string then the method returns false. If it is found, it moves onto the second character of the second string, picking up the comparison of string where the first character was found. \\
        \textbf{Python Signature:} & \texttt{def containsSubSequence(string1, string2):} \\
         \textbf{Unit Tests:} & \texttt{assert containsSubSequence("Qz\'UP3nQGY+yG|7", "Jf~vr") == False}\\
                              & \texttt{assert containsSubSequence("Qz\'UP3nQGY+yG|7", "Qz\'UP3nQGY+yG|7") == True}
    \end{tabularx}
    }
    \caption{Examples of automatically generated augmentation instances used for Unit Test-based RL training.}
    \label{fig:data}
\end{table*}
\clearpage

\begin{itemize}
    \item[] the lower rated one had 9. Removing outliers, the average number of lines of code is $6.8$ for preferred, and $6.9$ for dispreferred solutions. As such, there doesn't seem to be a clear bias of the critic's score wrt. solution length.
    \item The average of scores assigned to the highest and lowest rated solutions are extremely close -- $33.5377$ and $33.5375$. This indicates that distinguishing between ``good'' and ``bad'' correct solutions is indeed (unsurprisingly) a very hard problem for the critic.
    \item Certain variable names -- such as identifiers in for-loops -- seem more generic (i, j, k) in preferred solutions, and more specific (num, line) in dispreferred ones. This could be an artifact of such generic scoping variables being prevalent in the model's pretraining data.
    \item On average, preferred solutions seem more ``pythonic'' in nature. For example, we observed that constructs such as list comprehensions are used more frequently in preferred solutions. One could argue that this presence of language-specific built-ins is indeed indicative of ``better'' code than its absence.
\end{itemize}

Anecdotally, recalling our motivating example of implementations of the n-th Fibonacci number in Figures~\ref{fig:prompt} and~\ref{fig:alt} -- repeated in Figure~\ref{fig:fib_manual} for easier reference -- we find that the critic indeed agrees with our intuition that we would favour the latter; however, similar to the observation about average high/low scores above, the actual assigned scores for these two solutions are extremely close (identical to 4 significant figures). As such, we cannot claim that the critic seems to truly distinguish between (intuitively) better and worse code. 

Comparing the valid solutions for this problem generated by our actual model, we find the highest-rated one (Figure~\ref{fig:fib_sols}, left, score $33.5377$) is -- according to our intuition -- indeed better than the lowest-rated one (Figure~\ref{fig:fib_sols}, right, score $33.5375$), as the former makes use of recursion rather than a loop. Interestingly, the scores of the fictive example solutions given in Figures~\ref{fig:prompt} and~\ref{fig:alt} both lie between these maximum and minimum scores. This establishes a relative ranking of the four solutions as Figure~\ref{fig:fib_sols} (left) > Figure~\ref{fig:fib_manual} (right) > Figure~\ref{fig:fib_manual} (left)~> Figure~\ref{fig:fib_sols} (right); an order that according to our intuition seems justifiable.

While these examples should be taken with a grain of salt, we might take them as \emph{weak} evidence that a critic trained to predict unit test outcomes may indeed be able to meaningfully distinguish between different working solutions to the same problem, which is impossible to assess based on Unit Test performance alone.

\section{Test and Signature Conversion}
\label{app:conv}

Figure~\ref{fig:conv} shows a diagram of converting a Java code signature and Unit Tests into Python equivalents.

Our data augmentation method yields instances suitable for Unit Test-based RL training, consisting of a function description in natural language, as well as a Python signature and Python-style test cases, converted from Java as outlined above. We show a few examples of generated augmented instances in Table~\ref{fig:data}.

\end{document}